\documentclass[10pt,twocolumn,letterpaper]{article}

\usepackage[pagenumbers]{cvpr} 

\usepackage{graphicx}
\usepackage{amsmath}
\usepackage{amssymb}
\usepackage{booktabs}
\usepackage{multicol}
\usepackage{multirow}
\usepackage{pifont}

\usepackage[pagebackref,breaklinks,colorlinks]{hyperref}

\usepackage[capitalize]{cleveref}
\crefname{section}{Sec.}{Secs.}
\Crefname{section}{Section}{Sections}
\Crefname{table}{Table}{Tables}
\crefname{table}{Tab.}{Tabs.}

\def\cvprPaperID{*****} 
\def\confName{CVPR}
\def\confYear{2022}

\begin{document}

\title{Deep Incomplete Multi-view Clustering with Cross-view Partial Sample and Prototype Alignment}

\author{
{ Jiaqi Jin$^1$,
  Siwei Wang$^1$,
  Zhibin Dong$^1$,
  Xinwang Liu$^{1,}\thanks{Corresponding author}$ ~,
  En Zhu$^{1,*}$}\\
  $^1$School of Computer, National University of Defense Technology, Changsha, China\\
  \texttt{\{jinjiaqi,\,wangsiwei13,\,dzb20,\,xinwangliu,\,enzhu\}@nudt.edu.cn}
} 
\maketitle

\begin{abstract}
The success of existing multi-view clustering relies on the assumption of sample integrity across multiple views. However, in real-world scenarios, samples of multi-view are partially available due to data corruption or sensor failure, which leads to incomplete multi-view clustering study (IMVC). Although several attempts have been proposed to address IMVC, they suffer from the following drawbacks: i) Existing methods mainly adopt cross-view contrastive learning forcing the representations of each sample across views to be exactly the same, which might ignore view discrepancy and flexibility in representations; ii) Due to the absence of non-observed samples across multiple views, the obtained prototypes of clusters might be unaligned and biased, leading to incorrect fusion. To address the above issues, we propose a \textbf{C}ross-view \textbf{P}artial \textbf{S}ample and \textbf{P}rototype \textbf{A}lignment \textbf{N}etwork (CPSPAN) for Deep Incomplete Multi-view Clustering. Firstly, unlike existing contrastive-based methods, we adopt pair-observed data alignment as 'proxy supervised signals' to guide instance-to-instance correspondence construction among views. Then, regarding of the shifted prototypes in IMVC, we further propose a prototype alignment module to achieve incomplete distribution calibration across views. Extensive experimental results showcase the effectiveness of our proposed modules, attaining noteworthy performance improvements when compared to existing IMVC competitors on benchmark datasets.
\end{abstract}

\section{Introduction}\label{sec:intro}
In modern society, data collected for real-world applications usually stems from different domains, sensors or feature extractors, which gives rise to multi-view learning in literature\cite{yang2018multi, fu2020overview}. For instance, an autonomous car may have diverse sensors, and a movie is typically made up of images and audio. As an important paradigm of multi-view learning, multi-view clustering (MVC)\cite{xia2022tensorized, li2015large, pan2021multi, peng2019comic, ZJPACMMM,trosten2021reconsidering} divides data by exploiting the consistent and complementary information across multiple views. The success of existing multi-view clustering methods heavily relies on the fully-available data assumption. However, in practical applications, some views of instances are only partially available due to unstable sensors and damaged storage media. When some views are missing\cite{li2014partial}, the natural alignment property of same instances across multiple views is destroyed, which may result in insufficient mining of complementary and consistent information. To handle the incompleteness issue, many incomplete multi-view clustering algorithms (IMVC)\cite{wen2019unified, wang2022highly, liu2019multiple} with satisfactory performance have been proposed. Typical strategies are mainly based on matrix decomposition, incomplete multiple kernel learning and graph-based methods. Learning more discriminative consensus representations with incomplete view information is crucial to achieve better incomplete multi-view clustering performance. However, conventional IMVC methods are based on raw features and therefore, the performance heavily relies on the feature quality.

As deep neural networks\cite{lecun2015deep, hou2017deep, mateen2018fundus,DFCN2021,liuyue_HSAN,CCGC} have demonstrated superior performance in learning high-level representations, deep learning has become prevalent in various fields of computer vision and pattern classification. To this end, researchers have explored combining deep neural networks\cite{welling2016semi, velickovic2017graph} and conventional IMVC methods to improve clustering performance, and the resulting clustering method is called Deep Incomplete Multi-View Clustering (DIMVC) \cite{ijcai2020p447, wen2020dimc, xue2021clustering, wang2020icmsc, lin2022dual}. Most existing DIMVC methods adopt the principle of contrastive learning, treating different views of the same sample as positive pairs and their representations should be consistent. Such algorithms ignore the cross-view alignment correlation of samples and force instances of different views with unified representation, which may destroy the flexibility and variety of representations. We argue that the essence of IMVC task lies in discovering structural correspondence between different views, rather than rigidly and simply enforcing uniform representations across each view. In fact, IMVC can be regarded as a special case of '\textbf{partially-aligned}' multi-view setting, where the pair-observed data provides supervised instance-alignment signals.

\begin{figure}[t]
\begin{center}{
    \centering
    \includegraphics[width=0.48\textwidth]{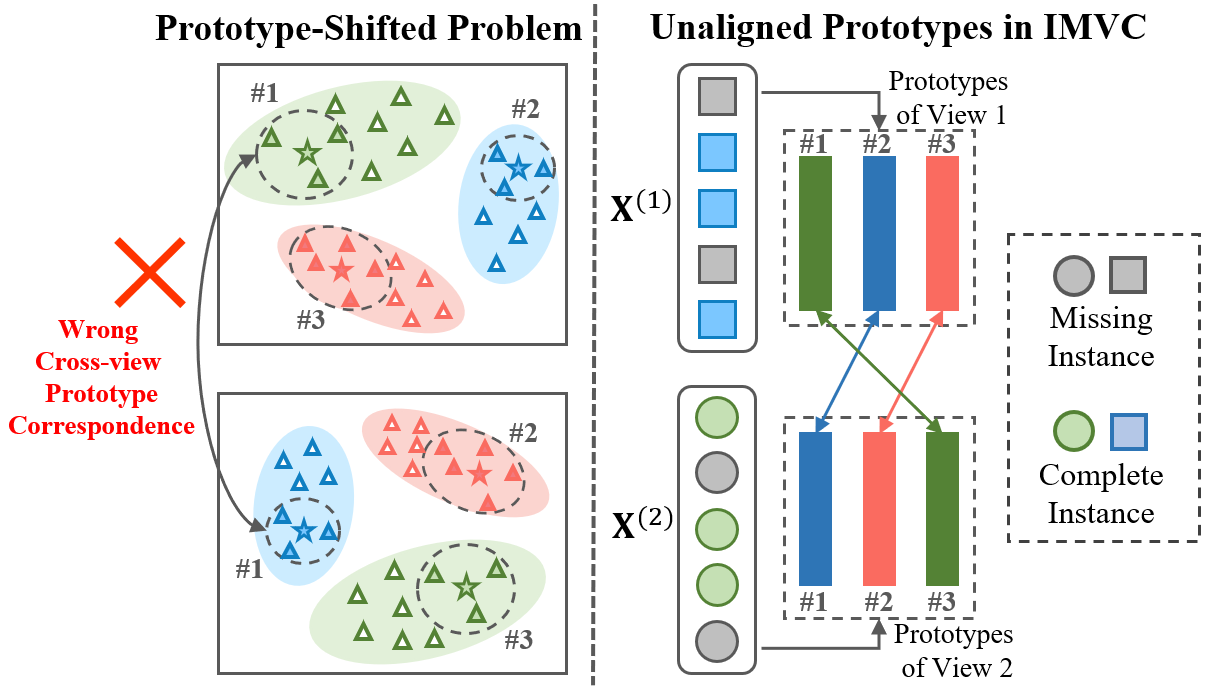}
    \caption{An example illustration of shifted prototype across multi-view caused by incomplete setting. With different missing status, the prototypes learned by incomplete multi-view data may be shifted and leads to wrong correspondences.}
    \label{PSP}}
\end{center}
\vspace{-20pt}
\end{figure}

Moreover, as shown in the Fig. \ref{PSP}, the distribution learned from the incomplete multi-view data can be biased due to inadequate multi-view data. Specifically, during the clustering task, flexible representations may cause prototypes of each cluster to shift and become biased, which we refer to as the Prototype-Shifted Problem (termed PSP). Such a problem has been demonstrated in the Anchor-Unaligned Problem \cite{wang2022align} for complete multi-view data, and undoubtedly has more essential impact on incomplete multi-view data. At the same time, contrastive-based DIMVC methods neglect this issue and do not explore relationships among different instances within the same view, which may further aggravate PSP. Therefore, it is necessary to match the relationship between the prototypes among views and perform clustering task accordingly.

\begin{figure*}[!htbp]
    \centering
    \includegraphics[width=1\textwidth]{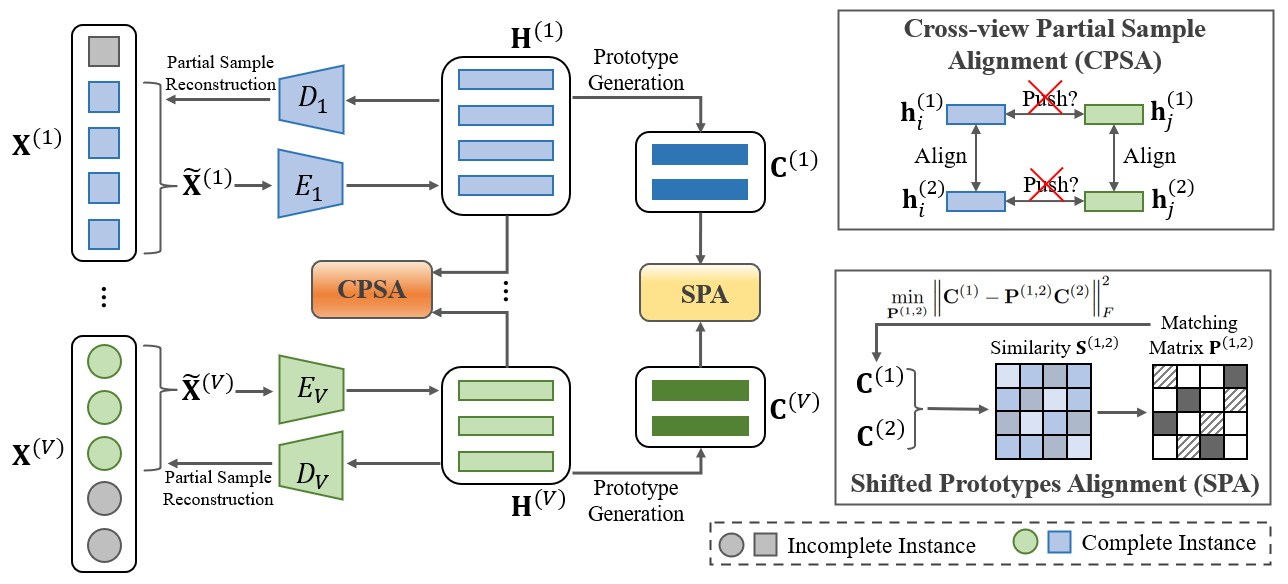}
    \caption{The framework of our proposed CPSPAN. CPSPAN consists of three modules: incomplete multi-view representation learning module, cross-view partial sample alignment module (CPSA) and shifted prototype alignment (SPA) module. Specifically, CPSA performs view-to-view alignment on learned representations of pair-observed instances. SPA explores the optimal matching correspondence between prototypes so as to align them. Finally, we use the structure embedding imputation strategy to fill in the missing embedding. Ultimately, we concatenate the complete and filled embedding and then perform kmeans to get the final result.}
    \label{fig:frame}
\end{figure*}

To address the aforementioned issues, we propose a novel approach termed \textbf{C}ross-view \textbf{P}artial \textbf{S}ample and \textbf{P}rototype \textbf{A}lignment \textbf{N}etwork (CPSPAN) for Deep Incomplete Multi-view Clustering to perform cross-view partial sample alignment and solve the prototype-shifted problem. The framework of CPSPAN is illustrated in Fig. \ref{fig:frame}. In detail, different from the contrastive learning mode, the cross-view instance alignment module establishes the view-to-view correspondence of samples through the pair-observed data in Fig. \ref{paireddata} between each pair-wise views, so as to mine the structural information between views. Afterwards, to address the prototype-shifted problem in incomplete scenario, the prototype alignment module takes one view's prototype set as anchors, and solves the permutation matrix between the two sets of prototypes, thereby establishing prototype-to-prototype correspondence based on optimal transport theory. Since prototypes are obtained based on samples, this module not only calibrates correspondences between cross-view shifted prototypes but also encodes the relationships between within-view samples. Ultimately, since our model is imputation-free upfront, in order to align the embeddings between views before finally performing feature fusion and clustering, we build cross-view structural relationship transfer for missing item imputations.

We summarize the major contributions of our work as follows,
\begin{itemize}
    \item We propose a novel deep network to handle IMVC task, termed as CPSPAN. Differ from existing multi-view contrastive learning manner, we considers the IMVC from a novel insight with \textbf{partially-aligned} setting. To this end, CPSPAN optimal maximizes matching alignment between paired-observed data and construct cross-view intersection.
    \item In order to solve the Prototype-Shifted Problem caused by incomplete information, CPSPAN proposes to further align the prototype sets between different views, so as to mine consistent cross-view structural information.
    \item  Extensive experiments have clearly demonstrated the effectiveness of the proposed cross-view partial sample and prototype alignment modules and the superiority over both conventional and deep SOTA methods.
\end{itemize}

\section{Related Work}\label{sec:related}
This section briefly describes the latest research progress of Incomplete Multi-View Clustering (IMVC) and Deep Incomplete Multi-View Clustering (DIMVC).

\subsection{Incomplete Multi-view Clustering}\label{related:sub1}
The incomplete multi-view clustering problem has attracted extensive attention of researchers in recent years. Existing methods for incomplete multi-view clustering mostly fall into three categories: (1) Matrix factorization based IMVC\cite{hu2019doubly, li2014partial}. Non-negative matrix factorization is dedicated to mining consensus representations of incomplete multi-view data. DAIMC\cite{hu2019doubly} is based on semi-NMF to obtain consensus information among multiple views. Partial multi-view clustering (PVC)\cite{li2014partial} exploits NMF to mine latent information of incomplete data.
(2) Kernel learning based IMVC\cite{liu2017optimal, liu2019multiple}. Kernel-based methods aim to recover the kernel matrix for incomplete multi-view data. ONKC\cite{liu2017optimal} enhances representability of the optimal kernel and strengthens negotiation between kernel learning and clustering. (3) Graph learning based IMVC\cite{zhang2020adaptive, zhao2016incomplete}. IMG\cite{zhao2016incomplete} extends PVC by exploring rich multi-view global structural information. Nevertheless, traditional IMVC methods have limitation in exploring the clustering-friendly representation of high-dimensional complex data, and they heavily rely on the quality of raw features.

\subsection{Deep Incomplete Multi-view Clustering}\label{related:sub2}
With the development of deep learning, researchers have realized the important role of representation learning in clustering task. The high-level feature mining capability of deep neural networks has attracted the attention of researchers, resulting in deep models being widely used in incomplete multi-view clustering tasks. The existing DIMVC methods can be divided into the following three  categories: (1) Autoencoders-based methods. By extracting feature of data, the deep autoencoder can learn the consistent representation to pad missing data and obtain impressive clustering performance. Typical methods are \cite{zhu2019multi}, \cite{wen2021structural}, \cite{wang2020icmsc} and \cite{lin2021completer}. (2) GANs-based methods. This kind of approaches explore the mutual representation between multiple views by generating adversarial networks and directly producing missing data\cite{wang2020generative, wang2021generative,wang2018partial,aggarwal2021generative, gui2021review}. (3) GCNs-based methods. GCNs-based methods explore common representations of multiple views through structural information among them\cite{wang2022graph, wang2022incomplete}. (4) Contrastive-based methods. The representative work is \cite{lin2022dual}. This method completes the missing data and explores the congruous representation through the comparison between multiple views. However, contrastive-based methods destroy the flexibility and diversity of representations and lose complementary information in incomplete multi-view clustering.

\begin{figure}[t]
\begin{center}{
    \centering
    \includegraphics[width=0.30\textwidth]{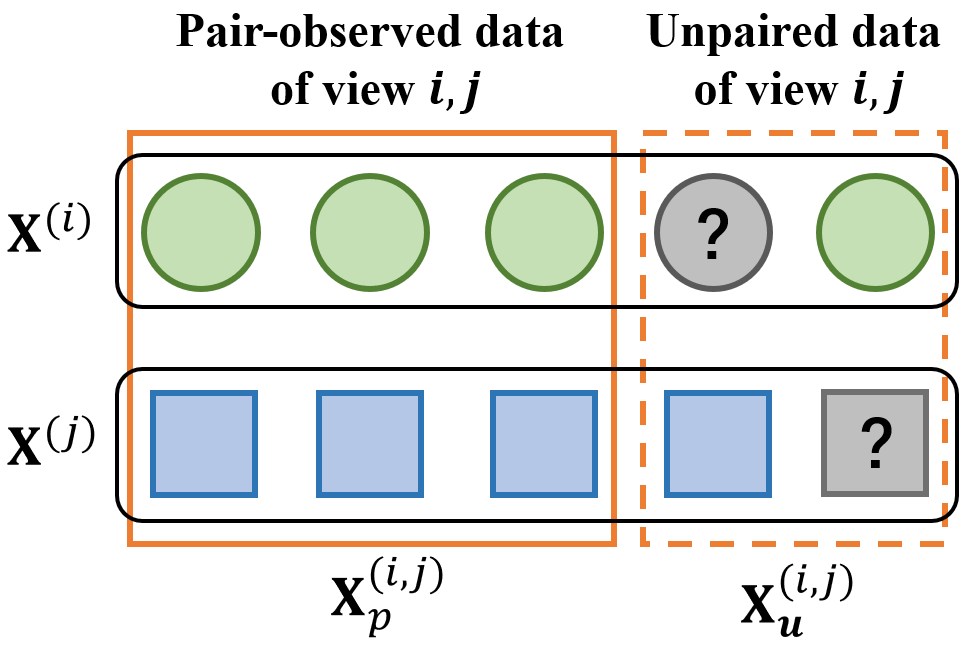}
    \caption{Definition of pair-observed data and unpaired data between view $i$ and $j$. Instances that are complete on both view $i$ and $j$ are referred as pair-observed data.}
    \label{paireddata}}
\end{center}
\end{figure}

\section{Proposed Method}\label{sec:method}
In this section, we first elaborate on the motivations of our work. Then we comprehensively introduce the proposed CPSPAN which consists of three jointly learning modules, namely, incomplete multi-view representation learning, cross-view partial sample alignment and shifted prototype alignment module.

\textbf{Notations}. Formally, a multi-view incomplete dataset $\{\mathbf{X}^{(v)} \in \mathbb{R}^{N \times D_v}\}_{v=1}^V$ consists of $N$ instances of $V$ views. We denote the samples with complete data as a set $\{\mathbf{{\tilde X}}^{(v)} \in \mathbb{R}^{N_v \times D_v}\}_{v=1}^V$, where $N_v$ denotes the number of complete instance in $v$-th view. As shown in Fig. \ref{paireddata}, we utilize $\mathbf{X}_p^{(i,j)}$ and $\mathbf{X}_u^{(i,j)}$ to denote the pair-observed and unpaired instances respectively. $K$ is the number of clusters.

\subsection{Representation Learning Module and Prototype Set Generation} 
\label{method:sub2}
\textbf{Incomplete Multi-view Representation Learning}.
In multi-view clustering tasks, deep autoencoders have been widely used to extract high-level representations of raw features\cite{guo2017improved, song2018self, xu2019ternary, zhang2020deep, lin2021completer}. Therefore, we equip each view with an auto-encoder to learn the clustering-friendly features of the respective view. To avoid inaccurate imputation or padding negatively impacting representation learning, we only train with the complete instance in each view. In order to avoid instance misalignment between views caused by missing data, we use resampling method to complement each view by sampling observed instances on each view. For the $v$-th view, an encoder-decoder pair, denoted as $E_v$ and $D_v$, is utilized to obtain the high-level embedding $\mathbf{H}^{(v)}$. Specifically, the feature extraction process of $E_v$ is expressed as $E_v(\mathbf{{\tilde X}}^{(v)}; \theta^v): \mathbf{{\tilde X}}^{(v)} \in \mathbb{R}^{N_v \times D_v} \to \mathbf{H}^{(v)} \in \mathbb{R}^{N_v \times d} $. The process of feature reconstruction of $\mathbf{H}^{(v)}$ by the decoder $D_v$ is formulated as $D_v(\mathbf{H}^{(v)}; \phi^v): \mathbf{H}^{(v)} \in \mathbb{R}^{N_v \times d} \to \hat{\mathbf{X}}^{(v)} \in \mathbb{R}^{N_v \times D_v} $, where $d$ is the dimension of the embedding on all views, $\theta^v$ and $\phi^v$ are the learnable parameters of autoencoder for the $v$-th view. Autoencoders are utilized to project the features of all views into the same dimensionality, which enables measuring correspondence between prototypes of different views in the same dimension. The reconstruction loss for all views is expressed as
\begin{equation}
\label{Lrec}
\begin{split}
\mathcal{L}_{rec} = &  \sum\limits_{v=1}^V { \left\|\mathbf{{\tilde X}}^{(v)} - D_v(\mathbf{H}^v)  \right\|_{F}^2} \\
= & \sum\limits_{v=1}^V \sum\limits_{i=1}^{N_v} { \left\|\mathbf{{\tilde x}}_i^{(v)} - D_v(E_v(\mathbf{{\tilde x}}_i^{(v)}))  \right\|_{2}^2},
\end{split}
\end{equation}
our model is first pre-trained in the above way, so far it is still independent on each view.

\textbf{Prototype Set Generation}. 
In the $d$-dimensional feature space of each view, the prototype set of each view $\left\{\mathbf{C}^{(v)} \right\}_{v=1}^V$ can be obtained by the following objective:
\begin{equation}
\label{getproto}
\min \limits_{\mathbf{C}^{(v)}} \sum\limits_{i=1}^{N_v} \sum\limits_{j=1}^K { \left\| \mathbf{h}_i^{(v)} - \mathbf{c}_j^{(v)} \right\|_{2}^2}
\end{equation}
where $\mathbf{C}^{(v)} \in \mathbb{R}^{K \times d}$. 
The subsequent cross-view partial sample alignment and shifted prototype alignment are performed between the features generated by the $E_v$ and $D_v$ of each view and the prototype set $\mathbf{C}^v$ generated by Eq. (\ref{getproto}).

\subsection{Cross-view Partial Sample Alignment}
\label{method:sub3}
In the context of missing data, the multi-view clustering method based on contrastive learning leverages $V$ representations of the same sample in different views as positive samples and views of other samples as negative samples. By doing so, the method forces the representations of different views corresponding to the same sample to converge, ultimately enabling the learning of consistent information across views. The commonly used loss function associated with this method is
\begin{equation}
\label{CLsmallloss}
\ell_{i}^{(v)}=- \sum_{\substack{r=1, \\ r\ne v}}^V \log \frac{\exp \left( h_i^{(v)} h_i^{(r)} / \tau \right)} {\sum_{\substack{t=1, \\ t\ne v}}^V \sum_{\substack{j=1, \\ j\ne i}}^N \exp \left( h_i^{(v)} h_j^{(t)} / \tau \right)},
\end{equation}
\begin{equation}
\label{CLloss}
\mathcal{L}_{cl} = \frac{1}{N} \sum_{v=1}^V \sum_{i=1}^N \ell_{i}^{(v)},
\end{equation}
where $h_i^{(r)}$ is the positive instance of the $i$-th sample in other view. $\tau$ is the temperature parameter to control the softness. After the optimization of the loss function in Eq. (\ref{CLloss}), the representation across views are forced to be exactly the same.

Different from previous methods based on the principle of contrastive learning, our cross-view instance alignment module can learn more flexible representations. The representations learned by our module are more flexible for two reasons. One is that the module uses the cosine similarity to measure the difference between paired data between two views in the feature space, rather than using the vector inner product or the Euclidean distance of the vector in the feature space, which is different from the clustering that pays more attention to mining the structural information of data. On the other hand, our method only restricts the representation of samples in different views to achieve the highest matching degree, and does not enforce the matching degree with other samples to be 0. This alignment approach avoids stretching samples that originally belong to the same cluster and is better suited to mining structural information.

Note that $\mathbf{X}_p^{(i,j)}$ represents the pair-observed data in view $i$ and $j$. The number of instances in $\mathbf{X}_p^{(i,j)}$ is $N_{(i,j)}$, and the corresponding paired embedding of $\mathbf{X}_p^{(i,j)}$ in the feature space is $\mathbf{H}_p^{(i,j)}$. We define the difference between the $p$-th instance in view $i$ and the $q$-th sample in view $j$ in $\mathbf{H}_p^{(i,j)}$ as 
\begin{equation}
\label{cossim}
\mathbf{S}_{p,q}^{(i,j)} = \frac{(\mathbf{h}_p^{(i,j)})^T \mathbf{h}_q^{(i,j)}}{\Vert\mathbf{h}_p^{(i,j)}\Vert \Vert\mathbf{h}_q^{(i,j)}\Vert},
\end{equation}
where $\mathbf{S}^{(i,j)} \in \mathbb{R}^{N_{(i,j)} \times N_{(i,j)}}$ is the similarity matrix between paired embeddings in view $i$ and $j$. 

As mentioned above, we only restrict all elements on the diagonal of $S^{(i,j)}$ to be 1, and do not impose constraints on other elements of $S^{(i,j)}$. We impose this constraint on paired embedding of each pair-wise views, so the loss function of our cross-view instance alignment module can be defined as
\begin{equation}
\label{loss2}
\mathcal{L}_{ia} = \sum\limits_{\substack{i=1,\cdots,V\\ j=i+1}} { \left\| diag(\mathbf{S}^{(i,j)}) - \mathbf{1}_{N_{(i,j)}} \right\|_{2}^2},
\end{equation}

\subsection{Shifted Prototype Alignment}
\label{method:sub4}
By matching the prototype-to-prototype correspondence between each pairwise views, this module calibrates the relationship between samples in a view and the relationship between prototypes on different views, thereby solving the prototype-shifted problem and further improving the clustering performance.

Formally, the crucial step of this module is to obtain the permutation matrix $\mathbf{P} \in \mathbb{R}^{K \times K}$ between prototypes of different views. To achieve prototype alignment, we optimize P as an integer linear programming problem with the objective of achieving an optimal match between two prototype sets. Formally,
\begin{equation}
\label{ot}
\begin{split}
 \mathop{\arg\min}_{\mathbf{P}} \; \; & Tr(\mathbf{D}\mathbf{P}^T), \\
   \text{s.t.} \;\; & P_{ij} \geq 0, {\forall} (i,j), \\
   & \mathbf{P}\mathbf{1}=\mathbf{1}, \\
   & \mathbf{P}^T \mathbf{1}=\mathbf{1},
 \end{split}
\end{equation}
where $\mathbf{P}$ is a square binary matrix that has exactly one entry of 1 in each row and column, and 0 elsewhere. We denote the permutation matrix between views $i$ and $j$ by $\mathbf{P}^{(i,j)}$. Taking the prototypes $\mathbf{C}^{(1)}$ and $\mathbf{C}^{(2)}$ on view 1 and view 2 as an example, if $\mathbf{C}^{(2)}$ is aligned to $\mathbf{C}^{(1)}$, after finding the best matching flow $\mathbf{P}^{(1,2)}$ between the two, use the permutation matrix $\mathbf{P}^{(1,2)}$ to left-multiply $\mathbf{C}^{(2)}$ to align to $\mathbf{C}^{(1)}$. The optimal formula for solving $\mathbf{P}^{(1,2)}$ is

\begin{equation}
\label{2view align}
\min \limits_{\mathbf{P}^{(1,2)}}{ \left\|\mathbf{C}^{(1)} - \mathbf{P}^{(1,2)} \mathbf{C}^{(2)} \right\|_{F}^2}.
\end{equation}

In order to make the optimization of the permutation matrix suitable for the derivation mechanism in deep neural networks, we follow the method in PVC\cite{huang2020partially} and use the differentiable substitution of the Hungarian algorithm to establish the correspondence of prototype sets. Formally,
\begin{equation}
\label{Hungarian1}
\Omega_1 = ReLU(\mathbf{P}^{(1,2)}),
\end{equation}

\begin{equation}
\label{Hungarian2}
\Omega_2 = \mathbf{P}^{(1,2)} - \frac{1}{n}(\mathbf{P}^{(1,2)}\mathbf{1}-\mathbf{1})\mathbf{1}^T,
\end{equation}

\begin{equation}
\label{Hungarian3}
\Omega_3 = \mathbf{P}^{(1,2)} - \frac{1}{n} \mathbf{1} (\mathbf{1}^T\mathbf{P}^{(1,2)} - \mathbf{1}^T),
\end{equation}
where $\Omega_1$, $\Omega_2$ and $\Omega_3$ project the permutation matrix $\mathbf{P}^{(1,2)}$ to the three constraint sets.

Following the above optimization process, we can obtain the permutation matrix and optimally match the prototype sets between each pairwise views, and the corresponding loss function is as follows
\begin{equation}
\label{loss3}
\mathcal{L}_{pa} = \sum\limits_{\substack{i=1,\cdots,V\\ j=i+1}} { \left\| \mathbf{C}^{(i)} - \mathbf{P}^{(i,j)} \mathbf{C}^{(j)} \right\|_{F}^2}.
\end{equation}

\subsection{Structure Embedding Imputation Strategy}
\label{method:sub5}
\begin{figure}[t]
\begin{center}{
    \centering
    \includegraphics[width=0.45\textwidth]{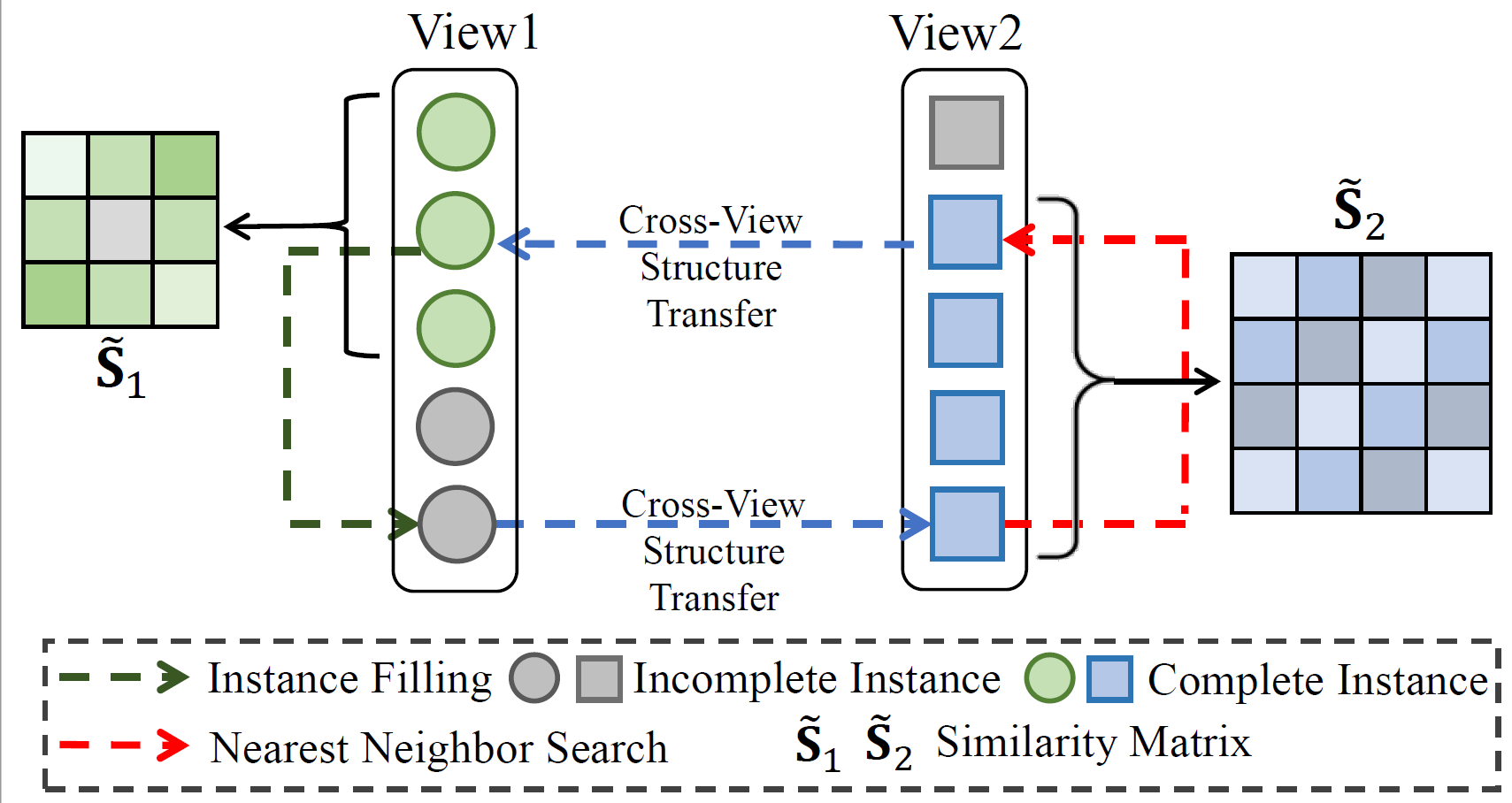}
    \caption{Visualization of Structure Embedding Imputation Strategy. We build cross-view structure transfer to impute instances with their close neighbours in latent space.}
    \label{1NN}}
\end{center}
\end{figure}

Since our model does not perform imputation during training, in order to align features between views, we propose a structure embedding imputation strategy to pad the missing embedding. Fig. \ref{1NN} is an example of our imputation strategy. Specifically, first we calculate the similarity matrix based on all complete embeddings in the two views respectively. For a missing feature in view 1, we find the nearest neighbor to its embedding in view 2, and then directly fill the missing feature with that neighbor's embedding in view 1.

\subsection{The Objective Function} 
\label{method:sub6}
With the above definitions, we propose the following objective function:
\begin{equation}
\label{totalloss}
\mathcal{L} = \mathcal{L}_{rec} + \alpha \mathcal{L}_{ia} + \beta \mathcal{L}_{pa},
\end{equation}
where $\mathcal{L}_{rec}$, $\mathcal{L}_{ia}$ and $\mathcal{L}_{pa}$ are within-view reconstruction loss for observable instances, cross-view partial sample alignment loss and shifted prototype alignment loss, respectively. In the experiments, we simply set both balance coefficients to 0.001.

\section{Experiments}\label{exper}

\subsection{Experimental Setting}\label{exper:sub1}
\noindent \textbf{Datasets and Evaluation Metrics}. We conduct experiments on five benchmark multi-view datasets, namely Caltech101-7, HandWritten, ALOI-100, YouTubeFace10 and EMNIST. Details of these data are shown in Table \ref{tab:dataset} below. In this experiment, we set the missing rate of the each dataset to [0.1, 0.3, 0.5, 0.7]. To evaluate the validity of the experiments, we used three widely used evaluation criteria including accuracy (ACC), normalized mutual information (NMI),and F-mea.
\begin{table}[!htbp]
\caption{Incomplete multi-view datasets in experiments.}
\label{tab:dataset}
\vspace{-8pt}
\centering
\resizebox{.49\textwidth}{!}{
\begin{tabular}{ccccc}
\toprule
Dataset         & Samples   & Clusters & Views & Dimensionality \\
\midrule
Caltech101-7    & 1400       & 7       & 5     & 1984/512/928/254/40  \\
HandWritten     & 2000       & 10      & 6     & 216/76/64/6/240/47   \\
ALOI-100        & 10800      & 100     & 4     & 77/13/64/125    \\
YouTubeFace10   & 38654      & 10      & 4     & 944/576/512/640  \\
EMNIST          & 280000     & 10      & 4     & 944/576/512/640  \\

\bottomrule
\end{tabular}}
\end{table}

\noindent \textbf{Baseline Methods}.
In experiments, our proposed algorithm is compared with seven state-of-the-art Incomplete multi-view clustering methods.
\textbf{Best Single-view Spectral Clustering
(BSV)}\cite{ng2001spectral} fills in missing data by mean on single view data. 
\textbf{PIC}\cite{liu2020efficient} strives to find a consensus feature matrix to complement incomplete multi-view data. \textbf{AWP}\cite{nie2018multiview} complements the missing data between multiple views through the structural information between multiple views. \textbf{CPM-Nets}\cite{zhang2019cpm} transforms the multi-view representation learning task into a degenerate process to achieve consistency and completion between multiple views.
\textbf{COMPLETER}\cite{lin2021completer} learns consistent representations through contrastive learning and maximizes the mutual information between multiple views.
\textbf{DCP}\cite{lin2022dual} learns consistent representations and complements information by maximizing mutual information and minimizing cross-entropy.
\textbf{DSIMVC}\cite{tang2022deep} dynamically complements the missing views from the learned semantic neighbors, solving the multi-view missing problem.

\noindent \textbf{Implementation}. Tne proposed CPSPAN is trained with the PyTorch platform. In the experiments, we set the pre-training batchsize to 256, the pre-training epoch to 200, and alignment epoch is 50. For all datasets, the autoencoders for all views are implemented by MLPs with the same structure, and the feature dimension in the embedding space is set to 10. The activation function is ReLU. We adopt Adam to optimize the deep model, the learning rate in the pre-training stage is set to 0.0005, and the learning rate in the alignment stage is set to 0.0001.

\begin{figure}[t]
\begin{center}{
    \centering
    \includegraphics[width=0.50\textwidth]{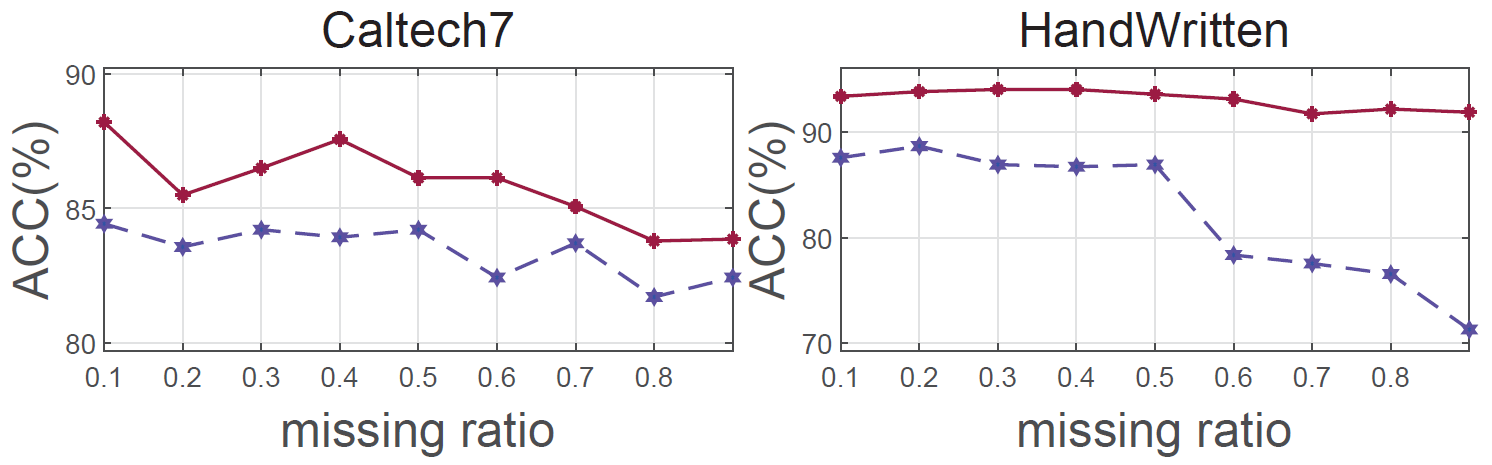}
    \caption{Flexibility verification of the cross-view instance alignment module. The red lines are the flexible features learned through our proposed CPSPAN, and the blue lines are the features learned after changing the loss function to contrastive loss.}
    \label{flexible_representation}}
\end{center}
\end{figure}

\subsection{Comparison with State-of-The-Arts}\label{exper:sub2}
\begin{table*}[!htbp]
\caption{The clustering performance comparisons on five benchmark datasets with varying missing rates. The best results are highlighted in bold, while the second best results are underlined. 'O/M' denotes the out-of-memory failure due to the size of the dataset.}
\vspace{0pt}
\label{tab:performance}
\centering
\resizebox{0.99\textwidth}{!}{
\begin{tabular}{|c|c|ccc|ccc|ccc|ccc|}
\hline
                            \multirow{2}{*}{}  & Missing rates & \multicolumn{3}{c|}{0.1}                                    & \multicolumn{3}{c|}{0.3}                                    & \multicolumn{3}{c|}{0.5}                                    & \multicolumn{3}{c|}{0.7}                                    \\ \cline{2-14}
                             & Metrics       & \multicolumn{1}{c}{ACC} & \multicolumn{1}{c}{NMI} & F-mea & \multicolumn{1}{c}{ACC} & \multicolumn{1}{c}{NMI} & F-mea & \multicolumn{1}{c}{ACC} & \multicolumn{1}{c}{NMI} & F-mea & \multicolumn{1}{c}{ACC} & \multicolumn{1}{c}{NMI} & F-mea \\ \hline
\multirow{8}{*}{\rotatebox{90}{\textbf{Caltech101-7}}}      & BSV          & 0.228 & 0.036 & 0.246 & 0.284 & 0.250 & 0.297 & 0.265 & 0.210 & 0.281 & 0.245 & 0.166 & 0.268 \\
                             & PIC   & 0.656 & 0.592 & 0.649 & 0.653 & 0.619 & 0.654 & 0.646 & 0.605 & 0.641 & 0.644 & 0.607 & 0.647 \\
                             & AWP    &0.779 & \underline{0.735} & 0.724 & 0.724 & \underline{0.677} & 0.677 & 0.661 & \underline{0.667} & 0.636 & \underline{0.772} & \underline{0.691} & \underline{0.695} \\
                             & CPM-Nets  & \underline{0.832} & 0.734 & \underline{0.811} & \underline{0.754} & 0.652 & \underline{0.721} & \underline{0.696} & 0.575 & \underline{0.693} & 0.611 & 0.499 & 0.605 \\
                             & COMPLETER  & 0.528 & 0.573 & 0.499 & 0.500 & 0.533 & 0.485 & 0.534 & 0.557 & 0.535 & 0.596 & 0.587 & 0.585 \\
                             & DCP  & 0.578 & 0.584 & 0.552 & 0.454 & 0.518 & 0.424 & 0.487 & 0.548 & 0.443 & 0.172 & 0.059 & 0.092 \\
                             & DSIMVC   & 0.649 & 0.565 & 0.642 & 0.629 & 0.537 & 0.622 & 0.554 & 0.471 & 0.549 & 0.419 & 0.343 & 0.391 \\
                             & \textbf{CPSPAN}   & \textbf{0.882} & \textbf{0.793} & \textbf{0.878} & \textbf{0.865} & \textbf{0.768} & \textbf{0.861} & \textbf{0.861} & \textbf{0.775} & \textbf{0.856} & \textbf{0.851} & \textbf{0.760} & \textbf{0.843} \\ \hline
\multirow{8}{*}{\rotatebox{90}{\textbf{HandWritten}}}   & BSV   &0.584 & 0.614 & 0.522 & 0.532 & 0.553 & 0.446 & 0.482 & 0.503 & 0.368 & 0.437 & 0.454 & 0.292 \\
                             & PIC   & 0.779 & 0.797 & 0.720 & 0.781 & 0.792 & 0.749 & 0.800 & 0.833 & 0.783 & 0.802 & 0.822 & 0.782 \\
                             & AWP   & 0.708 & 0.816 & 0.719 & \underline{0.867} & \underline{0.880} & 0.839 & \underline{0.860} & \underline{0.874} & \underline{0.834} & \underline{0.858} & \textbf{0.858} & \underline{0.823} \\
                             & CPM-Nets   & \underline{0.905} & 0.827 & \underline{0.905} & 0.840 & 0.754 & \underline{0.851} & 0.753 & 0.627 & 0.770 & 0.663 & 0.578 & 0.705 \\
                             & COMPLETER   &0.847 & \underline{0.852} & 0.833 & 0.610 & 0.679 & 0.615 & 0.481 & 0.614 & 0.485 & 0.730 & 0.730 & 0.727 \\
                             & DCP   & 0.682 & 0.793 & 0.644 & 0.769 & 0.788 & 0.765 & 0.670 & 0.738 & 0.652 & 0.708 & 0.742 & 0.686 \\
                             & DSIMVC   & 0.797 & 0.749 & 0.794 & 0.789 & 0.731 & 0.785 & 0.743 & 0.687 & 0.738 & 0.656 & 0.624 & 0.643 \\
                             & \textbf{CPSPAN}   & \textbf{0.934} & \textbf{0.880} & \textbf{0.933} & \textbf{0.940} & \textbf{0.888} & \textbf{0.940} & \textbf{0.936} & \textbf{0.879} & \textbf{0.935} & \textbf{0.917} & \underline{0.847} & \textbf{0.917} \\ \hline
\multirow{8}{*}{\rotatebox{90}{\textbf{ALOI-100}}}       & BSV   & 0.412 & \underline{0.741} & 0.268 & 0.369 & 0.741 & 0.237 & 0.322 & 0.740 & 0.236 & 0.290 & 0.742 & 0.238 \\
                             & PIC  & 0.698 & 0.608 & 0.436 & 0.633 & 0.345 & 0.242 & 0.541 & 0.465 & 0.365 & 0.351 & 0.355 & 0.235 \\
                             & AWP   & \underline{0.706} & 0.785 & \underline{0.579} & \underline{0.679} & \underline{0.785} & \underline{0.568} & \underline{0.674} & \underline{0.757} & \underline{0.538} & \underline{0.673} & \underline{0.757} & \underline{0.538} \\
                             & CPM-Nets   & 0.242 & 0.529 & 0.196 & 0.153 & 0.430 & 0.139 & 0.139 & 0.346 & 0.121 & 0.074 & 0.259 & 0.059 \\
                             & COMPLETER   & 0.169 & 0.421 & 0.143 & 0.174 & 0.419 & 0.164 & 0.144 & 0.384 & 0.135 & 0.143 & 0.403 & 0.173 \\
                             & DCP   & 0.272 & 0.521 & 0.255 & 0.230 & 0.477 & 0.220 & 0.227 & 0.488 & 0.221 & 0.213 & 0.451 & 0.215 \\
                             & DSIMVC   & 0.325 & 0.621 & 0.209 & 0.306 & 0.586 & 0.250 & 0.299 & 0.572 & 0.199 & 0.281 & 0.552 & 0.230 \\
                             & \textbf{CPSPAN}   & \textbf{0.712} & \textbf{0.864} & \textbf{0.738} & \textbf{0.688} & \textbf{0.856} & \textbf{0.654} & \textbf{0.709} & \textbf{0.864} & \textbf{0.679} & \textbf{0.686} & \textbf{0.859} & \textbf{0.662}  \\ \hline
\multirow{8}{*}{\rotatebox{90}{\textbf{YouTubeFace10}}}       & BSV   & O/M & O/M & O/M & O/M & O/M & O/M & O/M & O/M & O/M & O/M & O/M & O/M \\
                             & PIC   & O/M & O/M & O/M & O/M & O/M & O/M & O/M & O/M & O/M & O/M & O/M & O/M \\
                             & AWP   & O/M & O/M & O/M & O/M & O/M & O/M & O/M & O/M & O/M & O/M & O/M & O/M \\
                             & CPM-Nets   & \underline{0.772} & \underline{0.809} & \underline{0.753} & \underline{0.740} & \underline{0.768} & \underline{0.715} & \underline{0.698} & \underline{0.702} & \underline{0.673} & 0.657 & 0.621 & 0.633 \\
                             & COMPLETER   & 0.620 & 0.642 & 0.605 & 0.653 & 0.677 & 0.628 & 0.551 & 0.563 & 0.534 & 0.484 & 0.501 & 0.470\\
                             & DCP   & 0.746 & 0.782 & 0.725 & 0.659 & 0.644 & 0.617 & 0.604 & 0.577 & 0.586 & 0.628 & 0.604 & 0.572 \\
                             & DSIMVC   & 0.730 & 0.762 & 0.714 & 0.716 & 0.738 & 0.689 & 0.683 & 0.691 & 0.657 & \underline{0.664} & \underline{0.695} & \underline{0.642} \\
                             & \textbf{CPSPAN}   & \textbf{0.805} & \textbf{0.839} & \textbf{0.795} & \textbf{0.794} & \textbf{0.802} & \textbf{0.734} & \textbf{0.750} & \textbf{0.767} & \textbf{0.729} & \textbf{0.741} & \textbf{0.765} & \textbf{0.691} \\ \hline
\multirow{8}{*}{\rotatebox{90}{\textbf{EMNIST}}}       & BSV   & O/M & O/M & O/M & O/M & O/M & O/M & O/M & O/M & O/M & O/M & O/M & O/M  \\
                             & PIC   & O/M & O/M & O/M & O/M & O/M & O/M & O/M & O/M & O/M & O/M & O/M & O/M  \\
                             & AWP   & O/M & O/M & O/M & O/M & O/M & O/M & O/M & O/M & O/M & O/M & O/M & O/M  \\
                             & CPM-Nets   & 0.843 & 0.812 & \underline{0.837} & \underline{0.776} & \underline{0.749} & \underline{0.755} & \underline{0.720} & \underline{0.688} & \underline{0.692} & \underline{0.676} & \underline{0.621} & \underline{0.649} \\
                             & COMPLETER   & \underline{0.856} & \underline{0.827} & 0.835 & 0.762 & 0.704 & 0.711 & 0.650 & 0.597 & 0.618 & 0.533 & 0.485 & 0.509\\
                             & DCP   & 0.733 & 0.651 & 0.720 & 0.675 & 0.589 & 0.642 & 0.692 & 0.556 & 0.630 & 0.584 & 0.443 & 0.528 \\
                             & DSIMVC   & 0.710 & 0.672 & 0.694 & 0.639 & 0.572 & 0.597 & 0.585 & 0.524 & 0.537 & 0.530 & 0.472 & 0.508 \\
                             & \textbf{CPSPAN}   & \textbf{0.871} & \textbf{0.834} & \textbf{0.872} & \textbf{0.865} & \textbf{0.828} & \textbf{0.867} & \textbf{0.857} & \textbf{0.823} & \textbf{0.844} & \textbf{0.830} & \textbf{0.785} & \textbf{0.791} \\ \hline
\end{tabular}}
\end{table*}

In Table \ref{tab:performance} of the paper, our proposed algorithm is compared with seven state-of-the-art methods on three metrics. From the above chart, we have the following observations:
\begin{itemize}
\item Our proposed model significantly outperforms other incomplete multi-view clustering algorithms on all datasets in terms of all metrics. Under the 0.1 missing rate setting, for all the datasets, our proposed method are all significantly higher than the second best approaches, and outperform the second best by 5.0\%, 2.9\%, 0.6\%, 3.3\% and 1.5\%, respectively. This can demonstrate the superiority of our method on all datasets. Moreover, CPSPAN demonstrates greater stability as the missing rate varies. For instance, when the missing rate ranges from 0.1 to 0.7 on HandWritten, the performance only exhibits a modest 1.7\% decrease. Similarly, on the EMNIST, where the variation in the method's performance is greatest, the drop is only 6.4\%.
\item 
It can be seen that the CPSPAN can still achieve competitive effect with a high missing rate in the dataset. When the miss rate is 0.7, the proposed method is 7.9\%, 5.9\%, 1.3\%, 7.3\% and 15.4\% higher than the method with the second highest ACC in the five datasets, respectively. It can be seen that our proposed method can still achieve impressive results in the case of a relatively high missing rate, which demonstrates the stability of CPSPAN.
\end{itemize}
In conclusion, the above experiments have verified the effectiveness and stability of our proposed CPSPAN and it is pivotal to conduct pair-ovserved sample alignment and prototype alignment for IMVC task.

\subsection{Flexibility Validation of Cross-view Partial Sample Alignment} As mentioned in Section \ref{sec:intro}, our cross-view partial sample alignment strategy is more flexible than the commonly used contrastive-based methods, and can learn more clustering-friendly representation and mine better structural information of data. To verify this, we directly replace our pair-observed sample alignment loss with contrastive loss in Eq. (\ref{CLloss}). We validate our idea on both Caltech101-7 and HandWritten. From the results in Fig. \ref{flexible_representation}, it can be seen that our method outperforms contrastive-based approach at any missing rate, which indicates that our cross-view partial sample alignment module is more conducive to clustering representation.

\subsection{Stability Analysis of Imputation Strategy} 
We analyze the stability of the structure embedding imputation strategy by verifying it with the $i$-th nearest neighbor method, with $i$ ranging from 1 to 100. The results indicate that although the range of nearest neighbors varies widely, the performance only drops by less than 5\% on both datasets. This confirms the stability of our method. More detailed results are provided in the supplementary.

\begin{figure*}[h]
\begin{center}{
		\centering
			\subfloat[Raw Features]{{\includegraphics[width=0.25\textwidth]{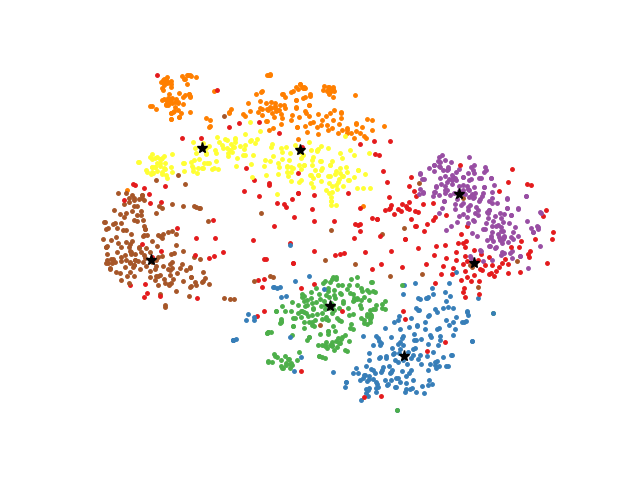}}}
			\subfloat[Final Result]{{\includegraphics[width=0.25\textwidth]{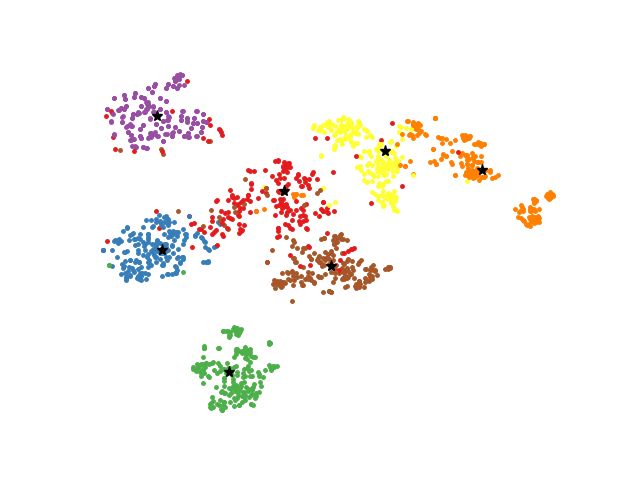}} }
   			\subfloat[Raw Features]{{\includegraphics[width=0.25\textwidth]{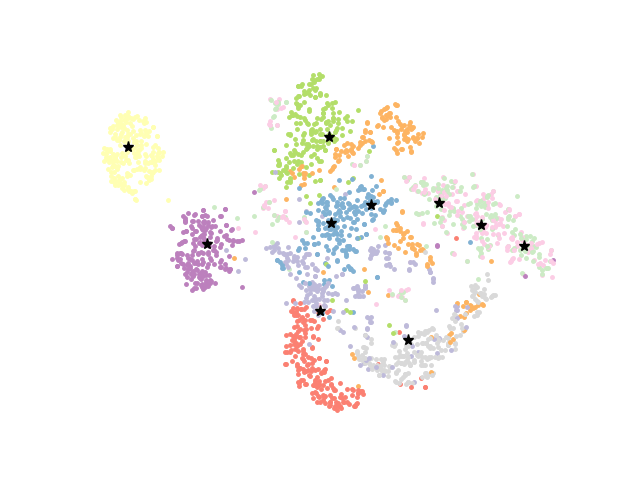}}}
			\subfloat[Final Result]{{\includegraphics[width=0.25\textwidth]{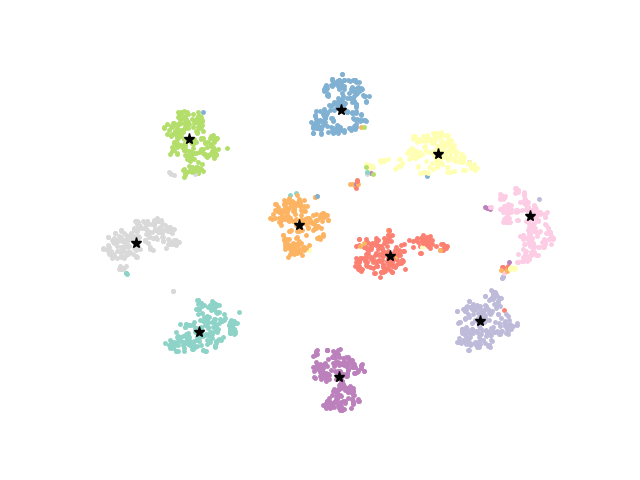}} }\\
			
			\caption{Visualization of Caltech101-7 and HandWritten. '*' indicates the prototype of each cluster.}
			\label{figure_EPOCH}
			}
\end{center}
\end{figure*}

\begin{table}[!htbp]
\caption{Ablation study on Caltech101-7. We select ACC as the evaluation metric. In the following table, \protect\ding{51} denotes CPSPAN with the component.}
\label{abaCaltech7}
\centering
\resizebox{0.43\textwidth}{!}{
\begin{tabular}{ccc|cccc}
\toprule[1pt]
\multicolumn{3}{c|}{Components} & \multicolumn{4}{c}{Missing Ratios} \\ \hline
$\mathcal{L}_{rec}$ & $\mathcal{L}_{ia}$ & $\mathcal{L}_{pa}$ & 0.1 & 0.3 & 0.5& 0.7 \\ \hline
\ding{51}   &               &               & 0.7050   & 0.6636   & 0.6286  & 0.5243   \\ 
\ding{51}   & \ding{51}  &               & 0.8429   & 0.8243   & 0.8057  & 0.8164   \\ 
\ding{51}   &               & \ding{51}  & 0.7757   & 0.7486   & 0.7071  & 0.7093   \\ 
\ding{51}   & \ding{51}  & \ding{51}  & \textbf{0.8821}  & \textbf{0.8650}   & \textbf{0.8614}  & \textbf{0.8507}   \\ 
\bottomrule[1pt]
\end{tabular}}
\end{table}

\begin{table}[!htbp]
\caption{Ablation study on HandWritten. We select ACC as the evaluation metric. In the following table, \protect\ding{51} denotes CPSPAN with the component.}
\label{ablationHW}
\centering
\resizebox{0.43\textwidth}{!}{
\begin{tabular}{ccc|cccc}
\toprule[1pt]
\multicolumn{3}{c|}{Components} & \multicolumn{4}{c}{Missing Ratios} \\ \hline
$\mathcal{L}_{rec}$ & $\mathcal{L}_{ia}$ & $\mathcal{L}_{pa}$ & 0.1 & 0.3 & 0.5& 0.7 \\ \hline
\ding{51}   &               &               & 0.8375   & 0.8115   & 0.8035  & 0.7780   \\ 
\ding{51}   & \ding{51}  &               & 0.9065   & 0.9035   & 0.8870  & 0.8600   \\ 
\ding{51}  &               & \ding{51}  & 0.8885   & 0.8600   & 0.8135  & 0.8075   \\ 
\ding{51}   & \ding{51}  & \ding{51}  & \textbf{0.9335 }  & \textbf{0.9400}   & \textbf{0.9355}  & \textbf{0.9170}   \\ 
\bottomrule[1pt]
\end{tabular}}
\end{table}

\begin{figure}[!htbp]

\begin{center}
{
\centering
\subfloat[\scriptsize{Caltech101-7}]{{\includegraphics[width=0.25\textwidth]{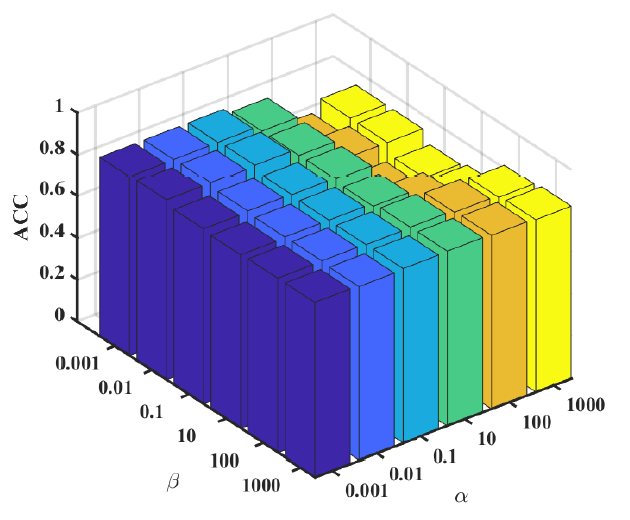}}}
\subfloat[\scriptsize{HandWritten}]{{\includegraphics[width=0.25\textwidth]{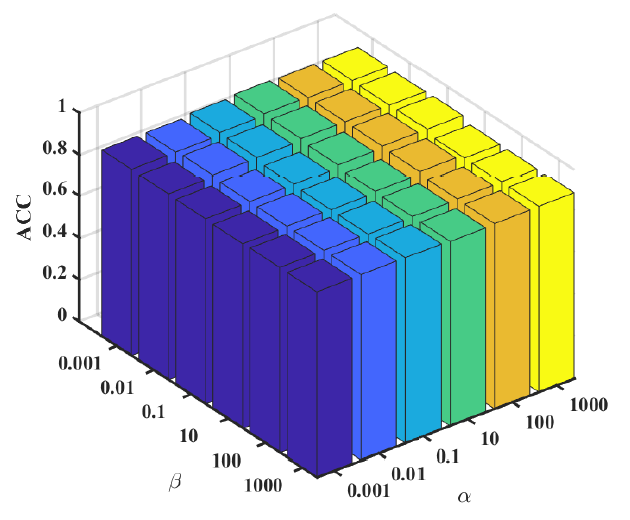}}}
\caption{Sensitivity analysis of $\alpha$ and $\beta$ for our method over HandWritten and Caltech101-7 (The missing ratio is 0.5).}
\label{sensitivityfig}
}
\end{center}
\label{ablation}
\end{figure}

\subsection{Parameter Sensitivity Analysis}
In order to analyze the effect of the two balance parameters $\alpha$ and $\beta$ in the loss function on the efficiency of the algorithm, we perform parameter sensitivity analysis on Caltech101-7 and HandWritten. We set the values of both parameters to [1e-3, 1e-2, 1e-1, 1e1, 1e2, 1e3]. From the results in the Fig. \ref{sensitivityfig}, we can find that $\beta$ has a greater impact on Caltech101-7, but both parameters have very little impact on HandWritten.

\subsection{Ablation Study on Modules}
The proposed CPSPAN contains three modules. To further verify the importance of each module, we conducted ablation experiments on Caltech101-7 and HandWritten datasets. The results are presented in Tables \ref{abaCaltech7} and \ref{ablationHW}, with missing rate of [0.1, 0.3, 0.5, 0.7]. Our findings indicate that both cross-view instance alignment and shifted prototype alignment modules contribute significantly to the improved performance of both datasets.

\subsection{Visualization} 
We visualize the clustering effect of CPSPAN on Caltech101-7 and HandWritten under the setting of 0.5 missing rate. Raw features and the final embedding are shown in Fig. \ref{figure_EPOCH} respectively. From the figure, we can clearly observe that after the training of our model, the instances from the same cluster become more compact, and the instances of different clusters are separated far away. In addition, the visualization results of the prototypes of each cluster further verify that after the prototype alignment module, the shifted prototype can be re-estimated and accurately calibrated.

\section{Conclusion}\label{conclu}

In this paper, we propose a novel Cross-view Partial Sample and Prototype Alignment framework for incomplete multi-view data termed CPSPAN. Different from other incomplete multi-view methods, CPSPAN uses pair-observed data  alignment to guide the correspondence reconstruction between samples. Aiming at the prototype-shifted problem in incomplete multi-view clustering, we also propose a shifted prototype alignment module to calibrate the prototype sets across views. Experiments demonstrate the superiority of our CPSPAN compared with conventional and deep IMVC SOTA methods.

\vspace{-5pt}
\section*{Acknowledgments}
\vspace{-5pt}
This work was supported by the National Key R\&D Program of China under Grant No. 2022ZD0209103 and the National Natural Science Foundation of China (project no. 62276271, 61906020, 61872371 and 62006237.

{\small
\bibliographystyle{ieee_fullname}
\bibliography{egbib}

\begin{thebibliography}{10}\itemsep=-1pt

\bibitem{aggarwal2021generative}
Alankrita Aggarwal, Mamta Mittal, and Gopi Battineni.
\newblock Generative adversarial network: An overview of theory and
  applications.
\newblock {\em International Journal of Information Management Data Insights},
  1(1):100004, 2021.

\bibitem{fu2020overview}
Lele Fu, Pengfei Lin, Athanasios~V Vasilakos, and Shiping Wang.
\newblock An overview of recent multi-view clustering.
\newblock {\em Neurocomputing}, 402:148--161, 2020.

\bibitem{gui2021review}
Jie Gui, Zhenan Sun, Yonggang Wen, Dacheng Tao, and Jieping Ye.
\newblock A review on generative adversarial networks: Algorithms, theory, and
  applications.
\newblock {\em IEEE Transactions on Knowledge and Data Engineering}, 2021.

\bibitem{guo2017improved}
Xifeng Guo, Long Gao, Xinwang Liu, and Jianping Yin.
\newblock Improved deep embedded clustering with local structure preservation.
\newblock In {\em Ijcai}, pages 1753--1759, 2017.

\bibitem{hou2017deep}
Xianxu Hou, Linlin Shen, Ke Sun, and Guoping Qiu.
\newblock Deep feature consistent variational autoencoder.
\newblock In {\em 2017 IEEE winter conference on applications of computer
  vision (WACV)}, pages 1133--1141. IEEE, 2017.

\bibitem{hu2019doubly}
Menglei Hu and Songcan Chen.
\newblock Doubly aligned incomplete multi-view clustering.
\newblock {\em arXiv preprint arXiv:1903.02785}, 2019.

\bibitem{huang2020partially}
Zhenyu Huang, Peng Hu, Joey~Tianyi Zhou, Jiancheng Lv, and Xi Peng.
\newblock Partially view-aligned clustering.
\newblock {\em Advances in Neural Information Processing Systems},
  33:2892--2902, 2020.

\bibitem{lecun2015deep}
Yann LeCun, Yoshua Bengio, and Geoffrey Hinton.
\newblock Deep learning.
\newblock {\em nature}, 521(7553):436--444, 2015.

\bibitem{li2014partial}
Shao-Yuan Li, Yuan Jiang, and Zhi-Hua Zhou.
\newblock Partial multi-view clustering.
\newblock In {\em Proceedings of the AAAI conference on artificial
  intelligence}, volume~28, 2014.

\bibitem{li2015large}
Yeqing Li, Feiping Nie, Heng Huang, and Junzhou Huang.
\newblock Large-scale multi-view spectral clustering via bipartite graph.
\newblock In {\em Twenty-ninth AAAI conference on artificial intelligence},
  2015.

\bibitem{lin2022dual}
Yijie Lin, Yuanbiao Gou, Xiaotian Liu, Jinfeng Bai, Jiancheng Lv, and Xi Peng.
\newblock Dual contrastive prediction for incomplete multi-view representation
  learning.
\newblock {\em IEEE Transactions on Pattern Analysis and Machine Intelligence},
  2022.

\bibitem{lin2021completer}
Yijie Lin, Yuanbiao Gou, Zitao Liu, Boyun Li, Jiancheng Lv, and Xi Peng.
\newblock Completer: Incomplete multi-view clustering via contrastive
  prediction.
\newblock In {\em Proceedings of the IEEE/CVF Conference on Computer Vision and
  Pattern Recognition}, pages 11174--11183, 2021.

\bibitem{liu2020efficient}
Xinwang Liu, Miaomiao Li, Chang Tang, Jingyuan Xia, Jian Xiong, Li Liu, Marius
  Kloft, and En Zhu.
\newblock Efficient and effective regularized incomplete multi-view clustering.
\newblock {\em IEEE transactions on pattern analysis and machine intelligence},
  43(8):2634--2646, 2020.

\bibitem{liu2017optimal}
Xinwang Liu, Sihang Zhou, Yueqing Wang, Miaomiao Li, Yong Dou, En Zhu, and
  Jianping Yin.
\newblock Optimal neighborhood kernel clustering with multiple kernels.
\newblock In {\em Proceedings of the AAAI conference on artificial
  intelligence}, volume~31, 2017.

\bibitem{liu2019multiple}
Xinwang Liu, Xinzhong Zhu, Miaomiao Li, Lei Wang, En Zhu, Tongliang Liu, Marius
  Kloft, Dinggang Shen, Jianping Yin, and Wen Gao.
\newblock Multiple kernel $ k $ k-means with incomplete kernels.
\newblock {\em IEEE transactions on pattern analysis and machine intelligence},
  42(5):1191--1204, 2019.

\bibitem{liuyue_HSAN}
Yue Liu, Xihong Yang, Sihang Zhou, Xinwang Liu, Zhen Wang, Ke Liang, Wenxuan
  Tu, Liang Li, Jingcan Duan, and Cancan Chen.
\newblock Hard sample aware network for contrastive deep graph clustering.
\newblock In {\em Proceedings of the AAAI conference on artificial
  intelligence}, 2023.

\bibitem{mateen2018fundus}
Muhammad Mateen, Junhao Wen, Sun Song, and Zhouping Huang.
\newblock Fundus image classification using vgg-19 architecture with pca and
  svd.
\newblock {\em Symmetry}, 11(1):1, 2018.

\bibitem{ng2001spectral}
Andrew Ng, Michael Jordan, and Yair Weiss.
\newblock On spectral clustering: Analysis and an algorithm.
\newblock {\em Advances in neural information processing systems}, 14, 2001.

\bibitem{nie2018multiview}
Feiping Nie, Lai Tian, and Xuelong Li.
\newblock Multiview clustering via adaptively weighted procrustes.
\newblock In {\em Proceedings of the 24th ACM SIGKDD international conference
  on knowledge discovery \& data mining}, pages 2022--2030, 2018.

\bibitem{pan2021multi}
Erlin Pan and Zhao Kang.
\newblock Multi-view contrastive graph clustering.
\newblock {\em Advances in neural information processing systems},
  34:2148--2159, 2021.

\bibitem{peng2019comic}
Xi Peng, Zhenyu Huang, Jiancheng Lv, Hongyuan Zhu, and Joey~Tianyi Zhou.
\newblock Comic: Multi-view clustering without parameter selection.
\newblock In {\em International conference on machine learning}, pages
  5092--5101. PMLR, 2019.

\bibitem{song2018self}
Jingkuan Song, Hanwang Zhang, Xiangpeng Li, Lianli Gao, Meng Wang, and Richang
  Hong.
\newblock Self-supervised video hashing with hierarchical binary auto-encoder.
\newblock {\em IEEE Transactions on Image Processing}, 27(7):3210--3221, 2018.

\bibitem{tang2022deep}
Huayi Tang and Yong Liu.
\newblock Deep safe incomplete multi-view clustering: Theorem and algorithm.
\newblock In {\em International Conference on Machine Learning}, pages
  21090--21110. PMLR, 2022.

\bibitem{trosten2021reconsidering}
Daniel~J Trosten, Sigurd Lokse, Robert Jenssen, and Michael Kampffmeyer.
\newblock Reconsidering representation alignment for multi-view clustering.
\newblock In {\em Proceedings of the IEEE/CVF conference on computer vision and
  pattern recognition}, pages 1255--1265, 2021.

\bibitem{DFCN2021}
Wenxuan Tu, Sihang Zhou, Xinwang Liu, Xifeng Guo, Zhiping Cai, En Zhu, and
  Jieren Cheng.
\newblock Deep fusion clustering network.
\newblock In {\em Proceedings of The Thirty-Fifth AAAI Conference on Artificial
  Intelligence}, pages 9978--9987, 2021.

\bibitem{velickovic2017graph}
Petar Velickovic, Guillem Cucurull, Arantxa Casanova, Adriana Romero, Pietro
  Lio, and Yoshua Bengio.
\newblock Graph attention networks.
\newblock {\em stat}, 1050:20, 2017.

\bibitem{wang2018partial}
Qianqian Wang, Zhengming Ding, Zhiqiang Tao, Quanxue Gao, and Yun Fu.
\newblock Partial multi-view clustering via consistent gan.
\newblock In {\em 2018 IEEE International Conference on Data Mining (ICDM)},
  pages 1290--1295. IEEE, 2018.

\bibitem{wang2020generative}
Qianqian Wang, Zhengming Ding, Zhiqiang Tao, Quanxue Gao, and Yun Fu.
\newblock Generative partial multi-view clustering.
\newblock {\em arXiv preprint arXiv:2003.13088}, 2020.

\bibitem{wang2021generative}
Qianqian Wang, Zhengming Ding, Zhiqiang Tao, Quanxue Gao, and Yun Fu.
\newblock Generative partial multi-view clustering with adaptive fusion and
  cycle consistency.
\newblock {\em IEEE Transactions on Image Processing}, 30:1771--1783, 2021.

\bibitem{wang2020icmsc}
Qianqian Wang, Huanhuan Lian, Gan Sun, Quanxue Gao, and Licheng Jiao.
\newblock Icmsc: Incomplete cross-modal subspace clustering.
\newblock {\em IEEE Transactions on Image Processing}, 30:305--317, 2020.

\bibitem{wang2022highly}
Siwei Wang, Xinwang Liu, Li Liu, Wenxuan Tu, Xinzhong Zhu, Jiyuan Liu, Sihang
  Zhou, and En Zhu.
\newblock Highly-efficient incomplete large-scale multi-view clustering with
  consensus bipartite graph.
\newblock In {\em Proceedings of the IEEE/CVF Conference on Computer Vision and
  Pattern Recognition}, pages 9776--9785, 2022.

\bibitem{wang2022align}
Siwei Wang, Xinwang Liu, Suyuan Liu, Jiaqi Jin, Wenxuan Tu, Xinzhong Zhu, and
  En Zhu.
\newblock Align then fusion: Generalized large-scale multi-view clustering with
  anchor matching correspondences.
\newblock {\em Advances in Neural Information Processing Systems}, 2022.

\bibitem{wang2022graph}
Yiming Wang, Dongxia Chang, Zhiqiang Fu, Jie Wen, and Yao Zhao.
\newblock Graph contrastive partial multi-view clustering.
\newblock {\em IEEE Transactions on Multimedia}, 2022.

\bibitem{wang2022incomplete}
Yiming Wang, Dongxia Chang, Zhiqiang Fu, Jie Wen, and Yao Zhao.
\newblock Incomplete multi-view clustering via cross-view relation transfer.
\newblock {\em IEEE Transactions on Circuits and Systems for Video Technology},
  2022.

\bibitem{welling2016semi}
Max Welling and Thomas~N Kipf.
\newblock Semi-supervised classification with graph convolutional networks.
\newblock In {\em J. International Conference on Learning Representations (ICLR
  2017)}, 2016.

\bibitem{wen2021structural}
Jie Wen, Zhihao Wu, Zheng Zhang, Lunke Fei, Bob Zhang, and Yong Xu.
\newblock Structural deep incomplete multi-view clustering network.
\newblock In {\em Proceedings of the 30th ACM International Conference on
  Information \& Knowledge Management}, pages 3538--3542, 2021.

\bibitem{ijcai2020p447}
Jie Wen, Zheng Zhang, and Yong Xu.
\newblock Cdimc-net: Cognitive deep incomplete multi-view clustering network.
\newblock In Christian Bessiere, editor, {\em Proceedings of the Twenty-Ninth
  International Joint Conference on Artificial Intelligence, {IJCAI-20}}, pages
  3230--3236. International Joint Conferences on Artificial Intelligence
  Organization, 7 2020.
\newblock Main track.

\bibitem{wen2019unified}
Jie Wen, Zheng Zhang, Yong Xu, Bob Zhang, Lunke Fei, and Hong Liu.
\newblock Unified embedding alignment with missing views inferring for
  incomplete multi-view clustering.
\newblock In {\em Proceedings of the AAAI conference on artificial
  intelligence}, volume~33, pages 5393--5400, 2019.

\bibitem{wen2020dimc}
Jie Wen, Zheng Zhang, Zhao Zhang, Zhihao Wu, Lunke Fei, Yong Xu, and Bob Zhang.
\newblock Dimc-net: Deep incomplete multi-view clustering network.
\newblock In {\em Proceedings of the 28th ACM international conference on
  multimedia}, pages 3753--3761, 2020.

\bibitem{xia2022tensorized}
Wei Xia, Quanxue Gao, Qianqian Wang, Xinbo Gao, Chris Ding, and Dacheng Tao.
\newblock Tensorized bipartite graph learning for multi-view clustering.
\newblock {\em IEEE Transactions on Pattern Analysis and Machine Intelligence},
  2022.

\bibitem{xu2019ternary}
Xing Xu, Huimin Lu, Jingkuan Song, Yang Yang, Heng~Tao Shen, and Xuelong Li.
\newblock Ternary adversarial networks with self-supervision for zero-shot
  cross-modal retrieval.
\newblock {\em IEEE transactions on cybernetics}, 50(6):2400--2413, 2019.

\bibitem{xue2021clustering}
Zhe Xue, Junping Du, Changwei Zheng, Jie Song, Wenqi Ren, and Meiyu Liang.
\newblock Clustering-induced adaptive structure enhancing network for
  incomplete multi-view data.
\newblock In {\em IJCAI}, pages 3235--3241, 2021.

\bibitem{CCGC}
Xihong Yang, Yue Liu, Sihang Zhou, Siwei Wang, Wenxuan Tu, Qun Zheng, Xinwang
  Liu, Liming Fang, and En Zhu.
\newblock Cluster-guided contrastive graph clustering network.
\newblock {\em arXiv preprint arXiv:2301.01098}, 2023.

\bibitem{yang2018multi}
Yan Yang and Hao Wang.
\newblock Multi-view clustering: A survey.
\newblock {\em Big Data Mining and Analytics}, 1(2):83--107, 2018.

\bibitem{zhang2020deep}
Changqing Zhang, Yajie Cui, Zongbo Han, Joey~Tianyi Zhou, Huazhu Fu, and
  Qinghua Hu.
\newblock Deep partial multi-view learning.
\newblock {\em IEEE transactions on pattern analysis and machine intelligence},
  2020.

\bibitem{zhang2019cpm}
Changqing Zhang, Zongbo Han, Huazhu Fu, Joey~Tianyi Zhou, Qinghua Hu, et~al.
\newblock Cpm-nets: Cross partial multi-view networks.
\newblock {\em Advances in Neural Information Processing Systems}, 32, 2019.

\bibitem{ZJPACMMM}
Junpu Zhang, Liang Li, Siwei Wang, Jiyuan Liu, Yue Liu, Xinwang Liu, and En
  Zhu.
\newblock Multiple kernel clustering with dual noise minimization.
\newblock In {\em Proceedings of the 30th ACM International Conference on
  Multimedia}, MM '22, page 3440–3450, New York, NY, USA, 2022. Association
  for Computing Machinery.

\bibitem{zhang2020adaptive}
Pei Zhang, Siwei Wang, Jingtao Hu, Zhen Cheng, Xifeng Guo, En Zhu, and Zhiping
  Cai.
\newblock Adaptive weighted graph fusion incomplete multi-view subspace
  clustering.
\newblock {\em Sensors}, 20(20):5755, 2020.

\bibitem{zhao2016incomplete}
Handong Zhao, Hongfu Liu, and Yun Fu.
\newblock Incomplete multi-modal visual data grouping.
\newblock In {\em IJCAI}, pages 2392--2398, 2016.

\bibitem{zhu2019multi}
Pengfei Zhu, Binyuan Hui, Changqing Zhang, Dawei Du, Longyin Wen, and Qinghua
  Hu.
\newblock Multi-view deep subspace clustering networks.
\newblock {\em arXiv preprint arXiv:1908.01978}, 2019.

\end{thebibliography}
}

\end{document}